\documentclass[journal]{IEEEtran}

\usepackage{caption}
\usepackage{cite}
\usepackage{amsmath}
\usepackage{algorithmic}
\usepackage{url}
\usepackage{graphicx}
\usepackage{array}
\usepackage{amssymb}
\usepackage{multirow}
\usepackage[caption=false,font=footnotesize]{subfig}
\usepackage[pagebackref=true,breaklinks=true,colorlinks,bookmarks=false]{hyperref}
\usepackage{amsthm}
\usepackage{dsfont}
\usepackage{bm}
\usepackage{booktabs}
\usepackage{xcolor}
\usepackage[T1]{fontenc}
\usepackage{algorithmic}
\usepackage[ruled, vlined]{algorithm2e}
\usepackage{dsfont}

\SetKwInput{KwInput}{Input}             
\SetKwInput{KwOutput}{Output}
\graphicspath{{fig/}}
\def\ie{\emph{i.e.}}

\def\al{\emph{et al. }}

\begin{document}

\title{FECANet: Boosting Few-Shot Semantic Segmentation with Feature-Enhanced Context-Aware Network}

\author{
	Huafeng~Liu$^{*}$,
	Pai~Peng$^{*}$,
	Tao~Chen,
	Qiong~Wang,
	Yazhou~Yao,
	and Xian-Sheng Hua

	\thanks{Huafeng Liu, Pai Peng, Tao Chen, Qiong Wang, and Yazhou Yao are with the School of Computer Science and Engineering, Nanjing University of Science and Technology, Nanjing, 210094, China.}
	\thanks{Xiansheng Hua is with the Terminus Group, Beijing, 100027, China.}	
	\thanks{*Equal contribution.}
}

% The paper headers
\markboth{}%
{Shell \MakeLowercase{\textit{et al.}}: SMCP}

\maketitle

\begin{abstract}
Few-shot semantic segmentation is the task of learning to locate each pixel of the novel class in the query image with only a few annotated support images. The current correlation-based methods construct pair-wise feature correlations to establish the many-to-many matching because the typical prototype-based approaches cannot learn fine-grained correspondence relations. However, the existing methods still suffer from the noise contained in naive correlations and the lack of context semantic information in correlations. To alleviate these problems mentioned above, we propose a Feature-Enhanced Context-Aware Network (FECANet). Specifically, a feature enhancement module is proposed to suppress the matching noise caused by inter-class local similarity and enhance the intra-class relevance in the naive correlation. In addition, we propose a novel correlation reconstruction module that encodes extra correspondence relations between foreground and background and multi-scale context semantic features, significantly boosting the encoder to capture a reliable matching pattern. Experiments on PASCAL-$5^i$ and COCO-$20^i$ datasets demonstrate that our proposed FECANet leads to remarkable improvement compared to previous state-of-the-arts, demonstrating its effectiveness. The source codes and models have been made available at \url{https://github.com/NUST-Machine-Intelligence-Laboratory/FECANET}.

\end{abstract}


% For peer review papers, you can put extra information on the cover page as needed:
\ifCLASSOPTIONpeerreview
	\begin{center} \bfseries EDICS Category: 3-BBND \end{center}
\fi
\IEEEpeerreviewmaketitle

\section{Introduction}

%%%%%%% --- Motivation, Challenge --- %%%%%%%

\IEEEPARstart{S}{emantic} segmentation\cite{wang2019learning,kang2018depth, zhan2019unmanned, chen2021semantically, yao2021non, chen2022saliency} is an essential component in modern computer vision, with many potential applications ranging from robotic manipulation  \cite{wong2017segicp} to medical image diagnosis \cite{asgari2021deep}. With the rapid development of convolutional neural networks \cite{zhang2018multilabel, he2016deep, szegedy2017inception, szegedy2015going, yao2021jo, sun2022pnp, sun2021webly}, fully supervised semantic image segmentation has made significant progress. However, the training of state-of-the-art semantic segmentation methods usually requires large-scale datasets \cite{deng2009large, lin2014microsoft, yao2017exploiting} with pixel-level annotation. It is challenging for them \cite{lin2017refinenet} to segment novel objects given very few annotated training images. Therefore, few-shot segmentation \cite{wang2019panet, nguyen2019feature, xie2021scale, zhang2020sg, wang2020few} is proposed to address the above novel class training issue, which aims to segment query images of unseen classes with only a handful of support images.

Currently, most few-shot segmentation methods \cite{zhang2019canet, yang2020prototype, liu2021anti, zhang2020sg, chen2021semantically} employ global average pooling over the foreground area of the support features to obtain the prototype vectors and utilize them to guide the segmentation of the query image.
As the compressed prototype vector retains only the most manifest information of the target class, these prototype-based methods simplify the many-to-many correspondence to a one-to-many matching problem. The lack of feature details hinders the prototype vector from conducting fine-grained matches with target objects in the query image. Therefore, solving many-to-many correspondence has enormous potential to explore better performance on few-shot semantic segmentation. HSNet \cite{min2021hypercorrelation} is a recently proposed many-to-many correspondence model that aims to learn visual correspondence by analyzing pattern relations in the 4D correlation space. It constructs many-to-many correlation tensors and processes them with center pivot 4D convolution. 

Although HSNet achieves superior performance for the few-shot segmentation task, there are still several shortcomings in its correlation construction. The main drawback is that HSNet directly uses the naive initial correlations generated by features from the backbone network as encoder input. The naive correlations may contain many noises caused by inter-class local similarities. These noises will mislead the encoder to learn inappropriate relation matching and segment background instances as objects of the target class. On the other hand, low relevance will be derived for regions of the same class due to intra-class diversity, which will result in incomplete object discovery. Another shortcoming of HSNet \cite{min2021hypercorrelation} is that hyper-correlation lacks context semantic information, which hinders the encoder from learning superior relation matching. Moreover, its constructed dense correlations filter background information in support images, which leads to the omission of many potential correspondence relations between foreground and background.\par

To alleviate these problems mentioned above, we propose a Feature-Enhanced Context-Aware Network (FECANet). First, inspired by the work of \cite{wang2018non, fu2019dual}, we propose a feature enhancement module (FEM) that employs a novel cross attention mechanism to facilitate accurate pattern matching between support and query features. Since directly leveraging features obtained by the backbone network will lead to naive correlations containing noise caused by intra-class diversity and inter-class similarity, we propose to enhance feature representation by suppressing local similarity between different classes and enhancing global similarity between the same classes. Specifically, we transform the self-attention mechanism of a single feature into the cross attention for paired features. We exchange information between query and support features to realize information communication by weighting support (query) features through the relevance of query (support) features to support (query) features. Moreover, our FEM is delicately designed with few trainable parameters to serve the purpose of class-agnostic segmentation and retain high generalization capability. To refine correlation construction, we explore the essence that is useful for the encoder to capture a reliable matching pattern. We notice that features endowed with global context are more robust to intra-class variations in CNN-based descriptors, which is beneficial for the encoder to segment objects accurately. Therefore, we design a correlation reconstruction module (CRM) that consists of dense integral correlation and global context correlation generations. Specifically, we keep background information in support features and aggregate diverse semantic features to generate a dense integral correlation that contains correspondence between foreground and background. In global context correlation generation, we employ a self-similarity module that encodes semantic information of local regions into a vector to generate a global context feature map. Furthermore, we leverage a multi-scale guidance module that fuses the global context semantic feature on multiple scales to capture more diverse and complex context features.

Our major contributions are summarized as follows: 
\begin{itemize}
\item We propose a feature enhancement module to filter noises in the correlation affected by inter-class similarity and intra-class diversity. It exchanges information between support and query features on spatial and channel dimensions to enhance feature representation.

\item We propose a correlation reconstruction module that encodes extra correspondence relations between foreground and background and multi-scale context semantic features, significantly boosting the encoder to capture a reliable matching pattern.

\item Extensive experiments on PASCAL-$5^i$ dataset and COCO-$20^i$ demonstrate state-of-the-art results compared to current methods.
\end{itemize}

The rest of this paper is organized as follows: 
the related work and preliminaries are described in Section \ref{related_work} and \ref{preliminaries} and our approach is introduced in Section \ref{approach}; we then report our evaluations on two widely-used datasets for few-shot image segmentation task in Section \ref{experiments}; we report the ablation studies in Section \ref{ab_study} and finally conclude our work in Section \ref{conclusion}.

\section{Related Work}
\label{related_work}

\subsection{Semantic Segmentation}
Semantic segmentation \cite{pei2022hierarchical, long2015fully} is a fundamental computer vision task that aims to classify each pixel in an image into specified object categories. Current methods are mainly based on fully convolutional networks (FCN) \cite{long2015fully}, which modifies existing classification architectures and replaces the final fully-connected layer with the convolution layer. FCN facilitates dense prediction and improves segmentation performance. In addition, encoder-decoder \cite{badrinarayanan2017segnet} is also a popular architecture to produce the segmentation map with high resolution. The encoder gradually down-samples the features to get abstract feature maps, and the decoder up-samples them to recover detailed information. Although traditional fully supervised segmentation methods achieve excellent performance, they extremely rely on datasets with a large number of pixel-level annotations \cite{everingham2015pascal}. They cannot generalize well when segmenting objects of unseen classes.

\subsection{Few-Shot Semantic Segmentation}
To resolve the generalization problem on scarcely annotated datasets, few-shot learning \cite{snell2017prototypical, vinyals2016matching, sung2018learning, finn2017model} methods are applied to the semantic segmentation field. Metric learning is one of the few-shot learning methods and has been widely used in few-shot segmentation tasks, e.g., learning knowledge using a prototype vector. Currently, most few-shot segmentation methods are mainly based on prototype vectors \cite{zhang2019canet, yang2020prototype, liu2021anti, zhang2020sg}. They use global average pooling over the foreground area of the support feature to get the global feature vector and compare spatial locations in the query feature with these feature vectors. Some work \cite{liu2020part, yang2020prototype, liu2021anti} attempt to add multiple prototype vectors that contain part-aware features of the target class to enrich support feature representations. Therefore, improved prototype vectors can gain diverse and fine-grained support object features. However, the above methods break the spatial structures of support features and leverage compressed prototype vectors to match all query feature pixels, leading to inaccurate predictions. Prototype representation only retains the most manifest features of support images, resulting in a lack of semantic information and detailed feature. To solve the above limitations, some few-shot semantic segmentation methods \cite{wang2020few, zhang2019pyramid} start to explore the direction of establishing element-to-element correspondence. They selectively extract useful information from the support image and keep spatial structure with detailed features. They construct relations between query and support features and perform semantic segmentation guided by relation aggregation. However, these methods only employ the graph attention to re-weight query feature maps by straightforward relationship guidance. Both different graph attention units (GAU) \cite{zhang2019pyramid} for graph reasoning and democratized graph attention (DGA) \cite{wang2020few} mechanisms on different intermediate features are insufficient to discover proper correspondence under the complex situation, such as large differences in object scales, large truncations, and occlusions. The recent method of HSNet \cite{min2021hypercorrelation} shows that utilizing correlation is the key to achieving advanced performance for few-shot segmentation, which resolves a many-to-many correspondence problem efficiently. Besides employing meta-learning frameworks, BAM \cite{lang2022learning} proposes to build an additional base learner branch to explicitly predict the regions of base classes in the query images. Then an ensemble module is adopted to facilitate the segmentation of novel objects via treating base classes as the regions that do not need to be segmented.

\begin{figure*}
\centering
\includegraphics[width=0.94\linewidth]{ 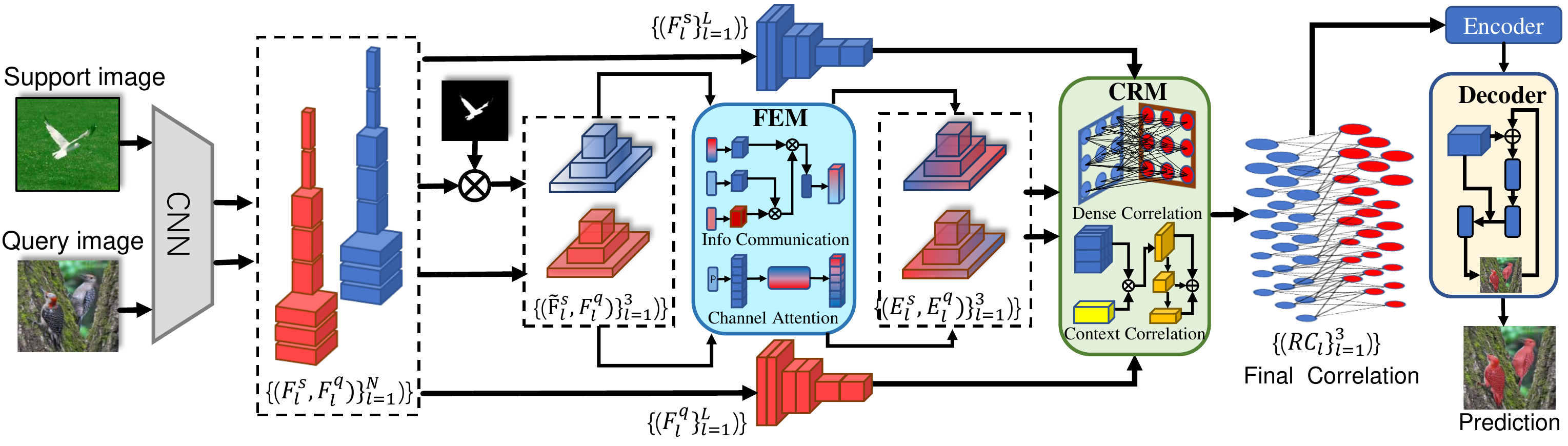}
\caption{Overall architecture of the proposed FECANet which consists of three main parts: feature enhancement module, correlation reconstruction module, and Residual 2D decoder. We refer the readers to Section \ref{approach} for details of the architecture.}
\label{fig:figue1}
\end{figure*}
\begin{figure*}
	\begin{minipage}{0.48\textwidth} 
		\centering
		\includegraphics[width=0.95\linewidth]{ 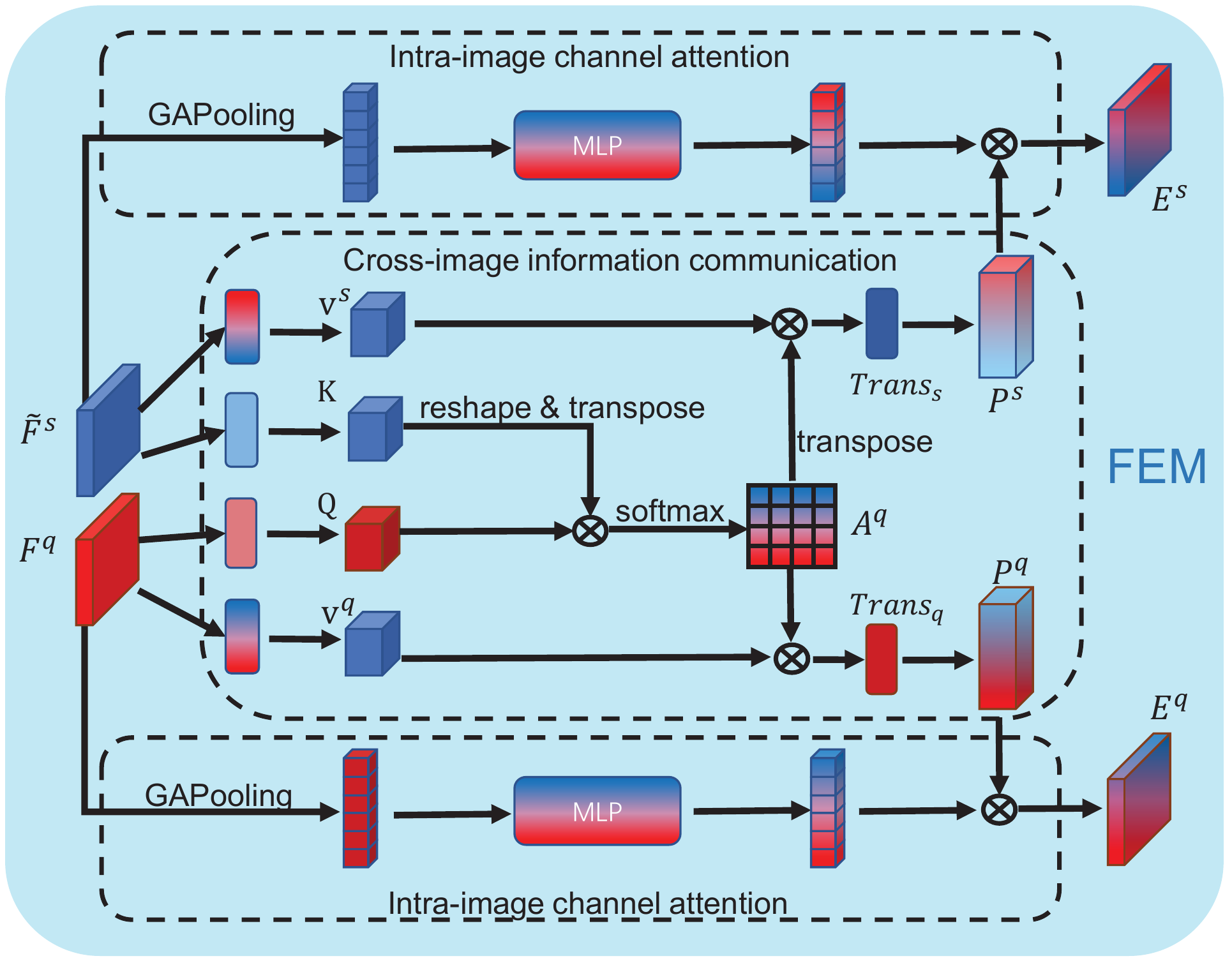}
		\caption{Implementation of feature enhancement module.} 
		\label{fig:figue2}
	\end{minipage}
	\hspace{0.2cm}
	\begin{minipage}{0.48\textwidth} 
		\centering
		\includegraphics[width=\linewidth]{ 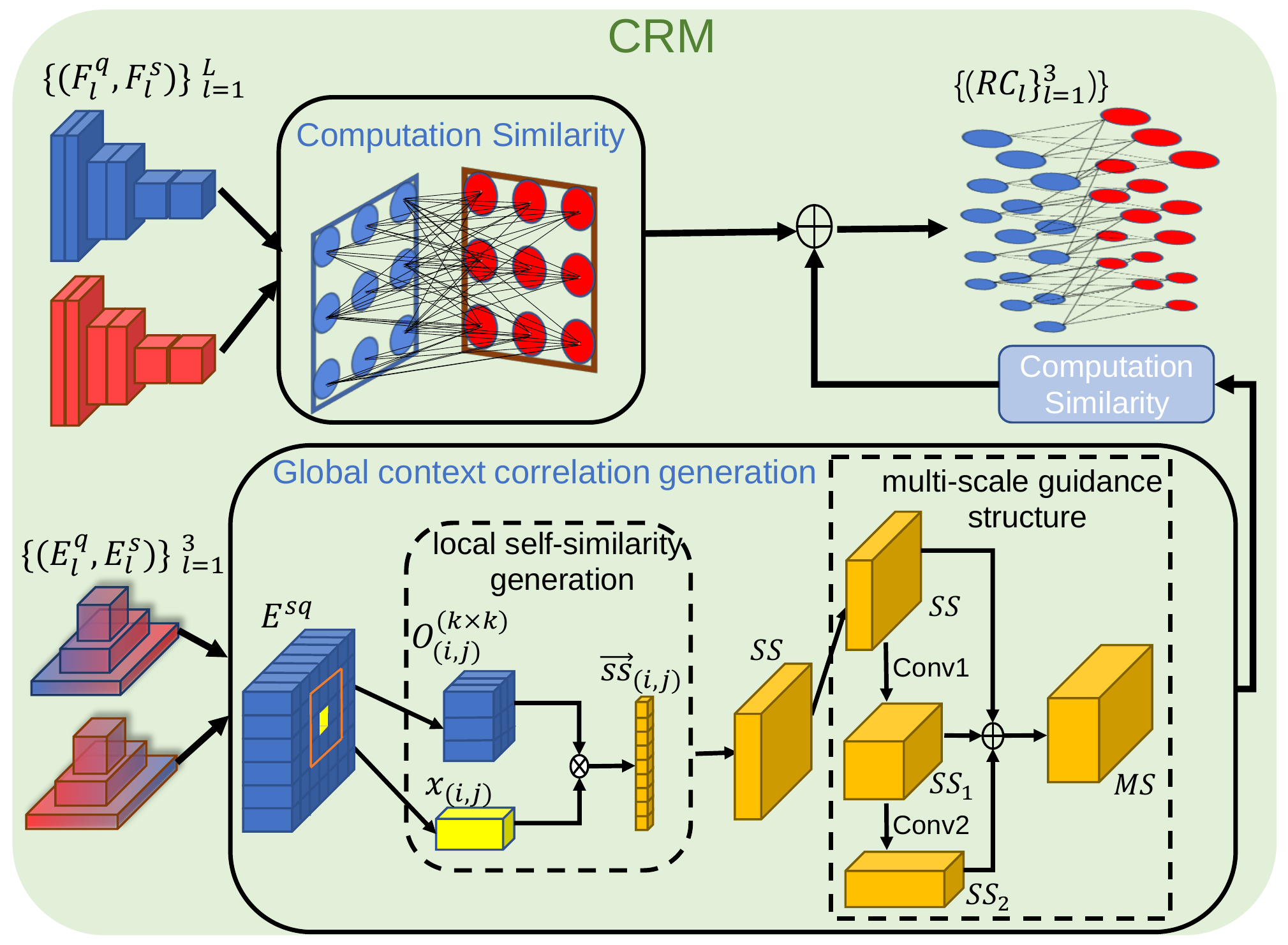}
		\caption{Illustration of correlation reconstruction module.}
		\label{fig:figue3}
	\end{minipage}
\end{figure*}

\subsection{Learning Visual Correspondence}
Finding visual correspondences \cite{balntas2017hpatches, ham2017proposal} across related images is a fundamental task in computer vision. It has seen a variety of applications in areas such as scene understanding \cite{liu2010sift}, object detection \cite{duchenne2011graph}, and semantic correspondence \cite{choy2016universal, kim2018recurrent,  kim2017fcss, rocco2017convolutional}. Recent works on semantic correspondence \cite{li2020correspondence, rocco2018neighbourhood} have achieved significant progress. They employ 4D convolutions on feature matching to find reliable semantic correspondences by analyzing local patterns in the 4D correlation feature space. Inspired by these works, HSNet \cite{min2021hypercorrelation} proposes a novel way to tackle the few-shot semantic segmentation problem. It learns semantic correspondences between support and query images by analyzing feature matching using an encoder consisting of center-pivot 4D convolution kernels. In addition, HSNet \cite{min2021hypercorrelation} constructs dense correlation using various intermediate CNN features as input to the encoder. Our work continues to focus on learning the visual correspondence way to tackle the few-shot segmentation problem. However, compared with previous work \cite{wang2018non, zhang2019canet, min2021hypercorrelation}, we refine naive initial correlations generated by features from the backbone network to enhance feature representation. In addition, we focus on constructing more effective correlation structures to promote the encoder to learn better correspondence relations between the query and the support images. 

\section{Preliminaries}
\label{preliminaries}
\subsection{Problem Setting}

The few-shot semantic segmentation task aims to perform segmentation on a scarce annotated dataset. We apply a popularly used meta-learning approach called episodic training to resolve the generalization problem caused by insufficient training data. Specifically, we divide overall data into a training set $D_{train}$ and test set $D_{test}$, and there is no overlap between the categories in $D_{train}$ and $D_{test}$. Both of train and test stages consist of several episodes. Each episode contains a support set $S$ and a query set $Q$, where $S = \{ x_i^s, m_i^s \}^k_{i=1} $ contains $k$-shot support images $x^s$ and corresponding binary masks $m^s$ for a certain class, and $Q = \{ x^q, m^q \} $ contains a query image $x^q$ to be segmented and related ground truth mask $m^q$. During the training phase, our model episodically collects data from $D_{train}$ and construct episode $ \{ \{ x_i^s, m_i^s \}^k_{i=1}, x^q$\} to learn mapping from predicted segmentation map to $m^q$. The training procedure is set to be aligned with the setting of evaluation. Once the model is trained, the parameter is fixed for evaluation on $D_{test}$ without further optimization.

\subsection{Revisit HSNet}
Inspired by the success of multi-level features and 4D convolutions in semantic correspondence tasks, Min \al combine the two influential techniques and design a framework, dubbed Hypercorrelation Squeeze Networks (HSNet) \cite{min2021hypercorrelation}, for the task of few-shot semantic segmentation. To construct hypercorrelations between a pair of input images with a rich set of correspondences, HSNet exploits diverse levels of feature representations from many different intermediate CNN layers. Specifically, after masking each support feature map with the support mask to discard irrelevant activations, a 4D correlation tensor is constructed using cosine similarity:
	\begin{equation}
		\hat{C}_l(x^q, x^s) = ReLU\left(\frac{F^q_l(x^q) \cdot \hat{F}^s_l(x^s)}{||F^q_l(x^q)|| \quad ||\hat{F}^s_l(x^s)||} \right). \label{eq_hs}
	\end{equation}
	Here, $x^q$ and $x^s$  denote 2-dimensional spatial positions of query feature map $F^q_l$ and the masked support feature map $\hat{F}^s_l$, respectively. $ReLU$ activation function is used to remove negative correlation scores. $\cdot$ denotes vector dot product. Then hypercorrelations for pyramidal layers are constructed by collecting these 4D tensors with concatenation along the channel dimension. 
	
	After obtaining hypercorrelation pyramid, HSNet leverage a 4D-convolutional pyramid encoder to squeeze it into a condensed feature map. Specifically, a squeezing block is adopted to squeeze the last two (support) spatial dimensions to a smaller size while maintaining the first two spatial (query) dimensions. Then the encoder leverages a mixing block to propagate relevant information to lower layers in a top-down fashion. After the iterative propagation, the last two (support) spatial dimensions are further compressed by average-pooling to provide a 2-dimensional feature map for the final decoder. To alleviate the computational burden caused by many 4D convolutions, center-pivot 4D convolution is devised with a weight-sparsification scheme to enable real-time inference.

\section{The Proposed Approach}
\label{approach}

%\subsection{Architecture Overview}
%\label{architecture_overview}

In this paper, we propose FECANet for tackling few-shot semantic segmentation. The framework is illustrated in Fig. \ref{fig:figue1}. We first leverage a convolutional neural network pre-trained on ImageNet as the backbone to extract a rich set of intermediate feature maps for both support and query images and denote them as dense pair-wise set $\{(F^s_l, F^q_l)\}^N_{l=1}$. Then, we select three pairs of support-query feature maps from the pair-wise set and filter out the background information in the support images with the support mask to obtain $ \{(\tilde{F^s_l}, F^q_l )\}^3_{l=1}$. Next, each pair of support and query features are fed into the proposed feature enhancement module (FEM) to enhance feature representation by exchanging information between query and support images. After that, the enhanced features $ \{(E^s_l, E^q_l) \}^3_{l=1}$ are conveyed into our proposed correlation reconstruction module (CRM), which captures context semantic information by a novel self-similarity module and establishes the local and global correlations between support and query features at each semantic level $\{RC_l\}_{l=1}^3$. Then, we apply a 4D convolution encoder to analyze correlations for capturing visual correspondence and fuse them from bottom to top layers. Finally, we acquire query representation by average pooling operation on the encoder output and deliver it to a residual 2D decoder to obtain the final query mask prediction.

\subsection{Feature Enhancement Module}

HSNet \cite{min2021hypercorrelation} directly uses the naive initial correlation between query and support features extracted by the backbone network, which suffers from the noise caused by the inter-class similarity and intra-class diversity. Inspired by the non-local block \cite{wang2018non} that encodes a broader range of contextual information to boost the representation capability, we develop a novel feature enhancement module to enhance feature representation by exchanging information between query and support features. Unlike the self-attention mechanism, our feature enhancement module aims to capture cross-image semantic similarities and differences between support and query features via information communication at the spatial level. Moreover, we notice that each channel map of features can be regarded as a semantic-specific response, and different semantic responses are associated with each other. Therefore, we design a channel attention module to re-weight the enhanced features for better inter-dependency refinement.\par

\textbf{Cross-image information communication.} As each pair of feature map is delivered to the module with different parameters but the same structure, we uniformity denote input feature pairs at any layer as $(\tilde{F^s_l}, F^q_l)$, where $\tilde{F^s} \in R^{C_l \times H_l \times W_l}$ and $F^q \in R^{C_l \times H_l \times W_l}$. We first leverage convolution layers on $\tilde{F^s}$ and $F^q$ to generate new feature maps $K$ and $Q$, where ${Q, K} \in R^{C_k \times H_l \times W_l}$ and $C_k$ is the channel number of low dimensional mapping space. Then we reshape them to the shape of $C_k \times N$, where $N = H_l \times W_l$ is the number of pixels. Different from self-attention, our query $Q$ and key $K$ features interact with each other for cross-image information communication. Therefore, as illustrated in Figure \ref{fig:figue2}, we design two branches to enhance support and query feature representation, respectively. After that, we perform a matrix multiplication between the transpose of $Q$ and $K$ to get the feature map and transpose this calculation result to obtain the feature map of another branch. Finally, we respectively apply a softmax layer on them to calculate the spatial attention map $A^q$ and $A^s$ for the query branch and support branch.

\begin{equation}
A^q_{ji} = \frac{exp(Q_i \cdot K_j)}{\sum_{i=1}^N exp(Q_i \cdot K_j)}, \quad
A^s = transpose(A^q).
\end{equation}
Here $A^q_{ji}$ measures the query $i^{th}$ position's impact on support $j^{th}$ position. The more similar feature representations of the two positions contribute to a greater correlation between them. \par

Meanwhile, we feed feature query $F^q$ and support $\tilde{F^s}$ feature into same convolution layer to generate two new feature map $V_q$ and $V_s$ respectively, where $ {V^q, V^s} \in R^{C_k \times H_l \times W_l} $ and $C_k$ is channel of low dimensional mapping space.Then we reshape them to the shape of $C_k \times N$. The common convolution layer is used to learn the knowledge mapping of similarity relation between support and query images. Then we perform a matrix multiplication between $V^q$ and $A^q$ and reshape the result to the shape of $ C_k \times H_l \times W_l$. For another branch, similar to the above calculation process, we perform a matrix multiplication between $V^s$ and $A^s$ and reshape the result to the shape of $ C_k \times H_l \times W_l$. Finally, we respectively feed them into two different transformation function layers $Trans_q$ and $Trans_s$ to generate result map $P^q$ and $P^s$. 

\begin{equation}
P^q =  Trans_q \Bigl(\sum_{i=1}^{N} (A^q_{ji} V^q_i) \Bigr), \quad
P^s =  Trans_s \Bigl(\sum_{i=1}^{N} (A^s_{ji} V^s_i) \Bigr).
\end{equation}
Here $Trans_q$ and $Trans_q$ are different transformation function layer consisted of convolution. The enhanced feature maps with the global contextual view can selectively aggregate contexts according to the spatial attention map. The similar semantic features achieve mutual gains, thus improving intra-class compact and semantic consistency.

\textbf{Intra-image channel attention.} After the cross-image information communication, we further leverage an intra-image channel attention module to enhance the support and query features representation. We first utilize a global pooling layer to aggregate the support/query features into a class-specific channel vector. Next, we employ a multi-layer perceptron consisting of fully connected layers and activation functions to map the support/query channel vector into a channel-wise feature space of similarity relation. The common multi-layer perceptron is used to improve the feature representation of specific semantics by exploiting the interdependencies in channel maps between support and query features. Then, we multiply the mapped channel vector with the position embedding features in a re-weighting manner to highlight the most important features selectively. Finally, we employ the residual connection to combine the output with the original input features. Our intra-image channel attention can be calculated as follows:
\begin{equation}
    E^s = Expand\bigl(MLP(Pooling(F^s))\bigr) \odot P^s + F^s,
\end{equation}

\begin{equation}
    E^q = Expand\bigl(MLP(Pooling(F^q))\bigr) \odot P^q + F^q.
\end{equation}
Here, $Expand$ function denotes vector extension operation that expands the vector to the spatial size of position embedding features, and $\odot$ denotes Hadamard product operation.

\subsection{Correlations Reconstruction Module}

Global context features provide robustness to intra-class variations in CNN-based descriptors. However, HSNet \cite{min2021hypercorrelation} lacks global context semantic information, which hinders the encoder from learning superior relation matching. Moreover, HSNet \cite{min2021hypercorrelation} entirely filters background information in the dense correlations with support mask maps, which hampers the encoder from learning potential relation matching from correspondence between foreground and background features to boost model segmentation performance.

In this section, we propose a novel correlation reconstruction module (CRM), where the collection of correlations is explicitly reconstructed locally and globally. The correlation reconstruction module consists of two parts: a \textit{Dense integral correlation generation} and a \textit{Global context correlation generation}. \par

\textbf{Dense integral correlation.} In this part, we select several intermediate features from the backbone network to construct a dense correlation collection. These features contain detailed information in support and query images, which boost the encoder's ability to learn fine-grained correspondence relations. Unlike hyper-correlation construction in HSNet, we abort to filter support feature with the object mask and keep background information in correlation. To be specific, we select $L$ pairs of deep features $\{(F^q_l, F^s_l)\}^L_{l=1}$ from a sequence of intermediate features maps in the backbone network. We directly compute similarity with each pair of query and support features at layer $l$ using cosine similarity:
\begin{equation}
C_l(x^q, x^s) = ReLU\left(\frac{F^q_l(x^q) \cdot F^s_l(x^s)}{||F^q_l(x^q)|| \quad ||F^s_l(x^s)||} \right). \label{eq_6}
\end{equation}
Here, $x^q$ and $x^s$ denote 2-dimensional spatial positions of query $F^q$ and support $F^s$ feature maps, respectively.   \par

Finally, we concatenate 4D similarity tensors having the same spatial sizes along channel dimension to construct correlation collection. We obtain three kinds of dense correlations in correlation collection and divide them into three different semantic layers. \par

\textbf{Global context correlation.} In this part, we introduce a novel local self-similarity method that encodes local spatial semantic features into context vector features to gain a global context feature map. Moreover, we introduce a multi-scale guidance structure to capture more global and complex context features. \par

\textit{Local self-similarity.} Inspired by previous works \cite{li2020correspondence, huang2019dynamic}, we compute the similarity of each spatial position within the local neighbor region of a specific size for self-similarity generation. Specifically, as show in Fig. \ref{fig:figue3}, we employ enhanced features $E^s_l$ and $E^q_l$ output from feature enhancement module as input, where $(E^s_l,E^q_l) \in  \{(E^s_l,E^q_l)\}^3_{l=1}$. As each pair of feature map is handled by the same process, we uniformity denote feature pair at any layer as $E^{sq} \in R^{C\times H \times W}$. To ensure that border location in feature map can be contained in local self-similarity computation, we apply zero-padding of size $(k-1)/2$ (we set k odd) on the feature map $E^{sq}$ to get padded feature map $E^{sq} \in R^{C \times (H + k - 1) \times (W + k - 1)}$. Then, for any position $x_{ij}$ in the padded feature map, where  $x_{(i,j)} \in R^{C\times 1 \times 1}$, we construct the local neighbor region of spatial size (k, k) centred at $x_{ij}$ as $O^{k \times k}_{(i,j)}$, where $O^{k \times k}_{(i,j)} \in R^{C \times k \times k}$. Then, we calculate dot product between $x_{ij}$ and every neighbor position in local region $O^{k \times k}_{(i,j)}$ to generate the self-similarity vector $\vec{ss}$, where $\vec{ss}_{(i,j)} \in R^{k^2 \times 1}$. Finally, the self-similarity features map $SS$ consists of self-similarity vector. The self-similarity features map is calculated as follows:

\begin{equation}
    t = (k-1) / 2,
\end{equation}
\begin{equation}
    \vec{ss}_{(i,j)} = [\tilde{x}_{(i-t, j-t)}^T \cdot x_{(i,j)}, \cdots, 
    \tilde{x}_{(i+t, j+t)}^T \cdot x_{(i,j)}],
\end{equation}

\begin{equation}
    SS = \{\vec{ss}_{(1,1)}, \cdots, \vec{ss}_{(H,W)}\}, \quad SS \in R^{k^2 \times H \times W}.
\end{equation}
Here, $x_{i,j}$ and $\tilde{x}_{i,j}$ denote the spatial position $(i, j)$ in feature map $E^{sq}$ and padded feature map $\tilde{E^{sq}}$ respectively. The blue {$k$} is the size of local neighbor region. $\vec{ss}_{(i,j)}$ denotes self-similarity vector generated at spatial position $(i, j)$ in feature map and $SS$ is a self-similarity feature map consisted of self-similarity vector $\vec{ss}$.

\textit{Multi-scale guidance.} As the size of the local region limits the channel number of the above self-similarity feature map, it is hard to contain a wider range of contextual semantic information. We leverage multi-scale self-similarities to further extract context semantic features and capture more complex self-similarity features. Our multi-scale similarity can provide a much larger local receptive field and richer contextual semantic information. Specifically, we refine the self-similarity output $SS$ with two successive convolutional layers $F_1$ and $F_2$ to generate $SS_1$ and $SS_2$. Then, we concatenate these different scale self-similarity features together with upsampling operations to form a multi-scale contextual feature map $ MS \in R^{C_{ss} \times H \times W}$:

\begin{equation}
    SS_1 = F_1(SS), \quad SS_2 = F_2(SS_1),
\end{equation}

\begin{equation}
    MS = SS \oplus upsampling(SS_1) \oplus upsampling(SS_2).
\end{equation}

Similar to Equation \eqref{eq_6}, global context correlation generation can be formulated as below:

\begin{equation}
GC_l(x^q, x^s) = ReLU\left(\frac{MS^q_l(x^q) \cdot MS^s_l(x^s)}
{||MS^q_l(x^q)|| \quad ||MS^s_l(x^q)||}\right)
\end{equation}
Here, $MS^q_l$ and $MS^s_l$ respectively denote multi-scale self-similarity derived from query and support enhanced feature map $E^q_l$ and $E^s_l$ at layer $l$, where $(E^q_l, \tilde{E^s_l}) \in \{(E^q_l, E^s_l)\}^3_{l=1}$. $x^s$ and $x^q$ denote 2-dimensional spatial positions of multi-scale support $MS^s_l$ and query $MS^q_l$ feature maps respectively. As the most helpful part for improving the results, our global context correlation optimizes the construction of the correlation. It leverages features endowed with global context to generate context correlation, which contains region-to-region correspondence. In CRM, our obtained region-to-region and pixel-to-pixel correlation can effectively promote the encoder to learn more comprehensive and deeper pattern-matching relationships. \par

Finally, we concatenate dense integral correlation $C_l$ and multi-scale global context correlations $GC_l$ along the channel dimension to construct collection. As the collection of correlation has different resolution sizes, we divide the collection into three groups $\{RC_l\}_{l=1}^3$. We then respectively deliver them into the three different 4D convolution encoders to analyze different semantic level correspondence relations between the query and the support images.

\begin{table*}[t]
\renewcommand\arraystretch{1.2}
\setlength{\tabcolsep}{1.7mm}
\caption{Performance on PASCAL-$5^i$\cite{shaban2017one} in mIoU and FB-IoU. }
\label{table:result_pascal}
\begin{center}
\begin{tabular}{c|lc|cccccc|cccccc}
\hline
\multirow{2}{*}{\textbf{Backbone}} & \multirow{2}{*}{\textbf{Methods}} &
\multirow{2}{*}{\textbf{Publication}} &
\multicolumn{6}{|c|}{\textbf{1-shot}} & \multicolumn{6}{c}{\textbf{5-shot}} \\ 
\multicolumn{1}{l|}{}      & \multicolumn{1}{l}{} & & ${5^0}$    & ${5^1}$    & ${5^2}$    & ${5^3}$    & mean & FB-IoU & ${5^0}$    & ${5^1}$   & ${5^2}$    & ${5^3}$    & mean & FB-IoU \\ 
\hline
\multirow{9}{*}{VGG16}      & OSLSM \cite{shaban2017one} & BMVC17               & 33.6 & 55.3 & 40.9 & 33.5 & 40.8 & 61.3   & 35.9 & 58.1 & 42.7 & 39.1 & 43.9 & 61.5   \\
                          & co-FCN \cite{rakelly2018conditional} & ICLRW18             & 36.7 & 50.6 & 44.9 & 32.4 & 41.4 & 60.1   & 37.5 & 50.0 & 44.1 & 33.9 & 41.4 & 60.2   \\
                          & AMP-2 \cite{siam2019adaptive}  & ICCV19             & 41.9 & 50.2 & 46.7 & 34.7 & 43.4 & 61.9   & 40.3 & 55.3 & 49.9 & 40.1 & 46.4 & 62.1   \\
                          & PANet \cite{wang2019panet}  &   ICCV19           & 42.3 & 58.0 & 51.1 & 41.2 & 48.1 & 66.5   & 51.8 & 64.6 & 59.8 & 46.5 & 55.7 & 70.7   \\
                          & PFENet \cite{tian2020prior} &  TPAMI20             & 56.9 & 68.2 & 54.4 & 52.4 & 58.0 & 72.0   & 59.0 & 69.1 & 54.8 & 52.9 & 59.0 & 72.3   \\
                          & HSNet \cite{min2021hypercorrelation} & ICCV21              & 59.6 & 65.7 & 59.6 & 54.0 & 59.7 & 73.4   & 64.9 & 69.0 & 64.1 & 58.6 & 64.1 & 76.6   \\
 
&NTRENet \cite{liu2022learning}& CVPR22& 57.7& 67.6&57.1&53.7&59.0&73.1&60.3&68.0&55.2&57.1&60.2&74.2\\  
&DPCN \cite{liu2022dynamic} &CVPR22&58.9&\textbf{69.1}&63.2&55.7&61.7&73.7&63.4&70.7&\textbf{68.1}&59.0&65.3&77.2\\                        
                           \cline{2-15}
                          & \textbf{Ours}   &      -        & \textbf{66.5} & 68.9 & \textbf{63.6} & \textbf{58.3} & \textbf{64.3} & \textbf{76.2}   & \textbf{68.6} & \textbf{70.8} & 66.7 & \textbf{60.7} & \textbf{66.7} & \textbf{77.6}   \\ \cline{1-15}
\multirow{13}{*}{ResNet50} & PANet \cite{wang2019panet} & ICCV19             & 44.0 & 57.5 & 50.8 & 44.0 & 49.1 & -      & 55.3 & 67.2 & 61.3 & 53.2 & 59.3 & -      \\
                          & PGNet \cite{zhang2019pyramid} &  ICCV19             & 56.0 & 66.9 & 50.6 & 50.4 & 56.0 & 69.9   & 57.7 & 68.7 & 52.9 & 54.6 & 58.5 & 70.5   \\
                          & PPNet \cite{liu2020part}  &  ECCV20           & 48.6 & 60.6 & 55.7 & 46.5 & 52.8 & 69.2   & 58.9 & 68.3 & 66.8 & 58.0 & 63.0 & 75.8   \\
                          & PFENet \cite{tian2020prior} & TPAMI20              & 61.7 & 69.5 & 55.4 & 56.3 & 60.8 & 73.3   & 63.1 & 70.7 & 55.8 & 57.9 & 61.9 & 73.9   \\
                          & RePRI \cite{boudiaf2021few}  & CVPR21            & 59.8 & 68.3 & 62.1 & 48.5 & 59.7 & -      & 64.6 & 71.4 & \textbf{71.1} & 59.3 & 66.6 & -      \\
                         & ASR\cite{liu2021anti} & CVPR21 & 55.2 & 70.3 & 53.3 & 53.6 & 58.1 & - & 58.3 & 71.8 & 56.8 & 55.7 & 60.9 & - \\
                         & HSNet \cite{min2021hypercorrelation} & ICCV21           & 64.3 & 70.7 & 60.3 & 60.5 & 64.0 & 76.7   & 70.3 & 73.2 & 67.4 & 67.1 & 69.5 & 80.6   \\
&CWT\cite{lu2021simpler}&ICCV21&56.3&62.0&59.9&47.2&56.4&-&61.3&68.5&68.5&56.6&63.7&-\\
&CMN \cite{xie2021few} &ICCV21&64.3&70.0&57.4&59.4&62.8&72.3&65.8&70.4&57.6&60.8&63.7&72.8\\
&SAGNN \cite{xie2021scale} &CVPR21&64.7&69.6&57.0&57.2&62.1&73.2&64.9&70.0&57.0&59.3&62.8&73.3\\
 &NTRENet \cite{liu2022learning}& CVPR22&65.4&72.3&59.4&59.8&64.2&77.0&66.2&72.8&61.7&62.2&65.7&78.4\\
 &ASNet \cite{kang2022integrative} &CVPR22&68.9&71.7&61.1&62.7&66.1&77.7&72.6&\textbf{74.3}&65.3&67.1&\textbf{70.8}&80.4\\
 &DPCN \cite{liu2022dynamic} &CVPR22&65.7&71.6&\textbf{69.1}&60.6&66.7&78.0&70.0&73.2&70.9&65.5&69.9&\textbf{80.7}\\

                         \cline{2-15}
                          & \textbf{Ours}  & - & \textbf{69.2} & \textbf{72.3} & 62.4 & \textbf{65.7} & \textbf{67.4} & \textbf{78.7}   & \textbf{72.9} & 74.0 & 65.2  & \textbf{67.8} & 70.0 & \textbf{80.7}  \\ \cline{1-15}
\end{tabular}
\end{center}
\end{table*}

\subsection{Residual 2D Decoder}

As we compress the 4D similarity tensor by average-pooling its last two (support) spatial dimensions, the 2D output of the encoder contains mixed context representation. To further refine the output of the 2D encoder, we develop the residual 2D decoder based on an ordinary 2D encoder consisting of a series of 2D convolutions and ReLU. Due to the appearance variances within the same category, the current support image might only guide the network to segment part of the object in the query image. Therefore, we construct a continuously-updated memory bank to store the prediction map of each query image to further refine the output of the 2D encoder, which provides rough object position clues for the next time of corresponding query image prediction. For each query image in the training stage, we fetch the corresponding query prediction map from the memory bank and concatenate it with the current encoder output. The initial query prediction map in the memory bank is padded with zeros. Then, the concatenated features are fed into the residual convolution module. The residual convolution module consists of a few 3x3 and 5x5 convolutions and employs a residual connection to fuse the input and output in each convolution operation. Next, the output from the residual convolution module is delivered to the convolution block to predict a two-channel mask map. Finally, we store the current prediction map of the query image in the memory bank for the next time of corresponding query image prediction.

After up-sampling the prediction map to the size of input query image, we define the segmentation loss as the cross-entropy between the prediction $P$ and query ground truth $M$:
	\begin{equation}
		L= - \sum_{h,w}\sum_{c\in C}M^{\left (h,w,c \right )}
		\log \left ( P^{\left ( h,w,c \right )} \right ) 
		\label{eq_fg}.
	\end{equation}
	Here, $C = \left \{ 0,1 \right \}$ is the class label that denotes whether the pixel belongs to the target class. $(h, w)$ is the size of the input query image and ground truth.

\begin{table*}[t]
\renewcommand\arraystretch{1.2}
\caption{Performance on COCO-$20^i$\cite{lin2014microsoft} in mIoU and FB-IoU.  }
\label{table:result_coco}
\begin{center}
\begin{tabular}{c|lc|cccccc|cccccc}
\hline
\multirow{2}{*}{\textbf{Backbone}} & \multirow{2}{*}{\textbf{Methods}} &
\multirow{2}{*}{\textbf{Publication}} &
\multicolumn{6}{|c|}{\textbf{1-shot}} & \multicolumn{6}{c}{\textbf{5-shot}} \\ 
\multicolumn{1}{l|}{}      & \multicolumn{1}{l}{} & & ${20^0}$    & ${20^1}$    & ${20^2}$    & ${20^3}$    & mean & FB-IoU & ${20^0}$    & ${20^1}$   & ${20^2}$    & ${20^3}$    & mean & FB-IoU \\ 
\hline
\multirow{3}{*}{VGG16}      
& FWB\cite{nguyen2019feature} & ICCV19 & 18.4 & 16.7 & 19.6 & 25.4 & 20.0 & - & 20.9 & 19.2 & 21.9 & 28.4 & 22.6 & - \\
& PRNet\cite{liu2020prototype} & CORR20 & 27.5 & 33.0 & 26.7 & 29.0 & 29.0 & - & 31.1 & 36.5 & 31.5 & 32.0 & 32.8 & - \\
%&SAGNN \cite{xie2021scale} &CVPR21&35.0&40.5&37.6&36.0&37.3&61.2&37.2&45.2&40.4&40.0&40.7&63.1\\
%&DPCN \cite{liu2022dynamic} &CVPR22&38.5&43.7&38.2&37.7&39.5&62.5&42.7&51.6&45.7&44.6&46.2&66.1\\
\cline{2-15}
& \textbf{Ours} & - & \textbf{34.1} & \textbf{37.5} & \textbf{35.8} & \textbf{34.1} & \textbf{35.4} & \textbf{65.5} &  \textbf{39.7} & \textbf{43.6} & \textbf{42.9} & \textbf{39.7} & \textbf{41.5} & \textbf{67.7} \\
\hline
                          
\multirow{12}{*}{ResNet50} 
& FWB\cite{nguyen2019feature} & ICCV19 & 16.9 & 17.9 & 20.9 & 28.8 & 21.1 & - & 19.1 & 21.4 & 23.9 & 30.0 & 23.6 & -\\
& PPNet\cite{liu2020part} &  ECCV20 & 28.1 & 30.8 & 29.5 & 27.7 & 29.0 & - & 39.0 & 40.8 & 37.1 & 37.3 & 38.5 & - \\
& PMM \cite{yang2020prototype} & ECCV20 & 29.3 & 34.8 & 27.1 & 27.3 & 29.6 & - & 33.0 & 40.6 & 30.3 & 33.3 & 34.3 & - \\
& RPMM \cite{yang2020prototype} & ECCV20 & 29.5 & 36.8 & 28.9 & 27.0 & 30.6 & - & 33.8 & 42.0 & 33.0 & 33.3 & 35.5 & - \\
& PFENet\cite{tian2020prior} & TPAMI20 & 36.5 & 38.6 & 34.5 & 33.8 & 35.8 & - & 36.5 & 43.3 & 37.8 & 38.4 & 39.0 & - \\
& REPRI\cite{boudiaf2021few} & CVPR21 & 32.0 & 38.7 & 32.7 & 33.1 & 34.1 & - & 39.3 & 45.4 & 39.7 & 41.8 & 41.6 & - \\

& ASR\cite{liu2021anti} & CVPR21 & 30.6 & 36.7 & 32.6 & 35.3 & 33.8 & - & 33.1 & 39.5 & 34.1 & 36.2 & 36.7 & - \\
& HSNet\cite{min2021hypercorrelation} & ICCV21 & 36.3 & 43.1 & 38.7 & 38.7 & 39.2 & 68.2 & 43.3 & 51.3 & 48.2 & 45.0 & 46.9 & 70.7 \\
&CWT \cite{lu2021simpler}&ICCV21&32.2&36.0&31.6&31.6&32.9&-&40.1&43.8&39.0&42.4&41.3&-\\
&CMN \cite{xie2021few}&ICCV21&37.9&\textbf{44.8}&38.7&35.6&39.3&61.7&42.0&50.5&41.0&38.9&43.1&63.3\\
&NTRENet \cite{liu2022learning} & CVPR22&36.8&42.6&39.9&37.9&39.3&68.5&38.2&44.1&40.4&38.4&40.3&69.2\\

\cline{2-15}
 & \textbf{Ours}  & - & \textbf{38.5} & 44.6 & \textbf{42.6} & \textbf{40.7} & \textbf{41.6} & \textbf{69.6}   & \textbf{44.6} & \textbf{51.5} & \textbf{48.4}  & \textbf{45.8} & \textbf{47.6} & \textbf{71.1}  \\ \cline{1-15} 
\end{tabular}
\end{center}
\end{table*}

\section{Experiments}
\label{experiments}

In this section, we first describe the experimental settings. Then we evaluate the proposed method compared with the recent state-of-the-art approaches.  \par

\subsection{Implement Details}
The backbone architecture employed in our few-shot segmentation model is ResNet \cite{he2016deep} and VGG \cite{he2016deep} pre-trained on ImageNet, e.g., ResNet50, VGG16. For ResNet, following HSNet \cite{min2021hypercorrelation}, we selectively extract features at the end of each bottleneck before ReLU activation: from $conv3_x$ to $conv5_x$ to construct dense integral correlation in CRM. We extract the last three-layer (\ie, $conv3$, $conv4$ and $conv5$) output as input to FEM for fair comparison with HSNet \cite{min2021hypercorrelation}. For VGG, we extract features after every pooling layer as input to FEM. We selectively extract features after every Conv layer in the last two building blocks: from $conv4_x$ to $conv5_x$ to construct dense integral correlation in CRM. We set spatial sizes of both support and query images to 400 $\times$ 400. The loss function is the mean of cross-entropy loss over the spatial location in the predicated segmentation map. The model is implemented in PyTorch \cite{paszke2019pytorch} and trained by the Adam optimizer with a batch size of 20 on one Nvidia Tesla V100 GPU. The learning rate is fixed to 1e-3 during training. We freeze the pre-trained backbone networks to prevent over-fitting. We use center-pivot 4D Conv kernel \cite{min2021hypercorrelation} to construct 4D convolution encoder.

\subsection{Datasets}
We evaluate our proposed method with experiments on two datasets: PASCAL-$5^i$\cite{shaban2017one}, COCO-$20^i$\cite{lin2014microsoft}. The PASCAL-$5^i$ dataset combines images from the PASCAL VOC 2012\cite{everingham2015pascal} and extra annotations from SDS \cite{hariharan2014simultaneous}. It contains 20 different object classes and evenly divide them into 4 folds: $\{5^i : i \in \{0,1,2,3\}\}$. The COCO-$20^i$ dataset is created for evaluation from a more challenging dataset MS COCO\cite{lin2014microsoft}. It contains 80 different object categories and evenly divided into 4 splits: $\{20^i : i \in \{0,1,2,3\}\}$. We conduct cross-validation over all the folds in the above two datasets. Specifically, for each fold $i$, we sample data from the other three folds to train the model. Then we randomly select data from  all images fold $i$ and evaluate them with the trained model.

\begin{table*}[t]
	\renewcommand\arraystretch{1.2}
	\setlength{\tabcolsep}{3.8mm}
	\caption{Ablation studies on $K$-shot setting. The results of other methods are from \cite{min2021hypercorrelation}.}
	\label{table:ab_shot}
	\begin{center}
		\begin{tabular}{c|cccc|cccc}
			\hline
			\multirow{2}{*}{\textbf{Method}} 
			& \multicolumn{4}{c}{PASCAL-$5^i$} & \multicolumn{4}{c}{COCO-$20^i$} \\ 
			&  { \textbf{1-shot} } &  { \textbf{5-shot} } &  { \textbf{10-shot} } &  { \textbf{20-shot}}  &  { \textbf{1-shot} } &  { \textbf{5-shot} } &  { \textbf{10-shot} } &  { \textbf{20-shot}}  \\ 
			\hline
			 {RPMM \cite{yang2020prototype}} &  {56.3} &  {57.3} &  {57.6} &  {-} &  {30.6} &  {35.5} &  {33.1} &  {-} \\
			 {PFENet\cite{tian2020prior}} &  {60.8} &  {61.9} &  {62.1} &  {-} &  {35.8} &  {39.0} &  {39.7} &  {- }\\
			 {REPRI\cite{boudiaf2021few}} &  {59.7} &  {66.6} &  {68.1} &  {-} &  {34.1} &  {41.6} &  {44.1} &  {-}\\
			 {HSNet\cite{min2021hypercorrelation}} &  {64.0} &  {69.5} &  {70.6} &  {-} &  {39.2} &  {46.9} &  {48.7} &  {-}\\
			 {\textbf{Ours}} & {\textbf{67.4  }} &  {\textbf{ 70.0  }} &  {\textbf{ 71.5  }} &  {\textbf{ 71.9 }}&  {\textbf{41.6  }} &  {\textbf{ 47.6  }} &  {\textbf{ 49.6   }} &  {\textbf{ 50.3  }}\\	
			\hline
		\end{tabular}
	\end{center}
\end{table*}

\subsection{Evaluation Metrics}
We take the commonly-used evaluation metrics, mean Intersection-over-Union (mIoU) and foreground-background IoU (FB-IoU), as our evaluation metrics. For each category, the IoU is calculated by $IoU = \frac{TP}{TP+FN+FP}$, where $TP$, $FN$, $FP$ respectively denote the number of true positive, false negative and false positive pixels of the predicted segmentation map. The metric of mIoU is defined as the average Intersection-over-Union (IoU) over all class: $ mIoU = \frac{1}{c} \sum_{c=1}^{C} IoU_c$, where $C$ is number of classes in the test fold and $c$ is index of relevant class. Compared with mIoU, FB-IoU ignores the difference among the object categories and regards all the categories as foreground class. Then FB-IoU computes mean of foreground and background IoUs over all test images: $FB$-$IoU = \frac{1}{2}IoU_F + \frac{1}{2}IoU_B$, where $IoU_F$ denote foreground IoU value and $IoU_B$ denote background IoU value. 

\subsection{$K$-shot Evaluation}
For the $K$-shot case ($K>1$), given $K$ support images with mask and the query image, we forward pass each support and query image into our model for $K$ times. Specifically, for each support image, we fetch the corresponding query prediction map from the memory bank and concatenate it with the current encoder output to segment query images in the residual 2D decoder. The initial query prediction map in the memory bank is padded with zeros. For each forward transmission in the test stages, we update the memory bank with the current predictive segmentation map of the query image. Finally, we sum up all the $K$ predictions and divide the output score by the maximum voting score. If their values are greater than the hyper-parameter threshold $\tau$ = 0.5, we assign foreground labels to the pixels otherwise background.

\subsection{Experimental Results}

We evaluate the proposed model on PASCAL-$5^i$ and COCO-$20^i$ and compare the results with recent methods.

\textbf{PASCAL-$5^i$.} Table. \ref{table:result_pascal} summarizes the mean-IoU and FB-IoU result on PASCAL-$5^i$ under 1-shot and 5-shot for all the compared methods. As can be seen, our approach achieves better results than other state-of-the-art methods for both VGG16 and ResNet backbones. Specifically, in the 1-shot setting, our method surpasses the state-of-the-art by 2.6\% and 0.7\% with VGG-16 and ResNet-50, respectively, demonstrating the proposed model's effectiveness in a one-shot setting. Moreover, our method also achieves competitive or better results than other methods with two different backbones in the 5-shot setting, demonstrating the proposed model's effectiveness in a multi-shot setting. In addition, our FB-IoU segmentation result sets a new state of the arts compared to other methods. 

\textbf{COCO-$20^i$.} Table. \ref{table:result_coco} summarizes 1-shot and 5-shot results on COCO-$20^i$ for all the compared methods. In the 1-shot setting, our method surpasses the state-of-the-art by 6.4\% and 2.3\% with VGG-16 and ResNet-50, respectively, verifying its superiority in few-shot segmentation task. Our method performs significantly better than other methods with two different backbones in the 5-shot setting, demonstrating the proposed model's effectiveness in a multi-shot setting.

 {Compared with PASCAL-$5^i$, the COCO-$20^i$ dataset contains more categories of objects and is much more difficult due to challenges like object scale difference and occlusion. Previous prototype-based approaches break the spatial structures of support features and leverage compressed prototype vectors to match all query feature pixels. Consequently, they cannot accurately match support and query images for objects with large-scale differences and occlusion. In contrast, our proposed method can learn more comprehensive and deeper semantic relationships by analyzing the correlation containing many-to-many matching. Therefore, our proposed method achieves much better results than baselines on COCO-$20^i$ compared with PASCAL-$5^i$.
}

 {
Table \ref{table:ab_shot} also presents the results of increasing the number of shots to 10 and 20. As we can see, the improvements are consistent on both PASCAL-$5^i$ and COCO-$20^i$ datasets, indicating the effectiveness of our approach.
}

\section{Ablation Studies}
\label{ab_study}

\begin{figure*}[t]
\centering
\includegraphics[height=8.8cm]{ 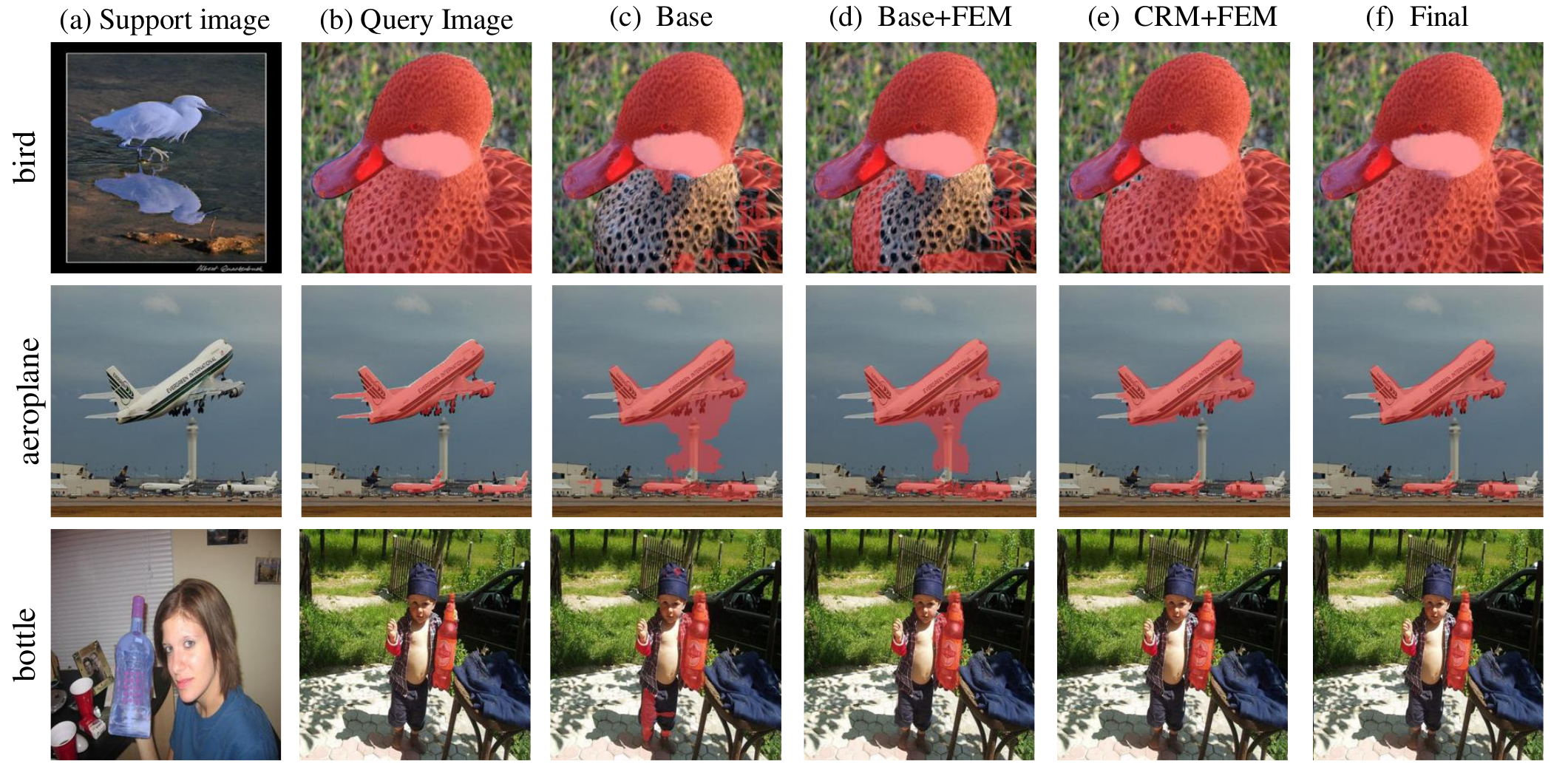}
\caption{Example result on PASCAL-$5^i$ dataset for component analysis. From left to right, we show that (a) support image with ground truth mask region in blue, (b) query image with ground truth mask region in red,  (c) base prediction, (d) base+FEM prediction (e) CRM+FEM
prediction, (f) final prediction (CRM+FEM+RD).}
\label{fig:ab_module}
\end{figure*}

\begin{table}[t]
\renewcommand\arraystretch{1.15}
\setlength{\tabcolsep}{4.5mm}
\caption{Ablation studies on the effect of different components. \textbf{Base:} naive correlation generated by three pairs of feature extracted by the backbone network, \textbf{FEM:} Feature Enhancement Module \textbf{CRM:} Correlation Reconstruction Module, \textbf{RD:} 2D Residual Decoder.}
\label{table:ab_module}
\begin{center}
\begin{tabular}{cccccc}
\hline
\textbf{Base} & \textbf{FEM} & \textbf{CRM} & \textbf{RD} & \textbf{Mean} \\
\hline
    $\surd$  &    &     &    & 61.3      \\
    $\surd$  &$\surd$    &     &    & 62.5      \\
     $ \surd$ & $\surd$     &  $\surd$  &      &  67.1  \\
     $\surd$ &  $\surd$    & $\surd$   &  $\surd$    & 67.4 \\
\hline
\end{tabular}
\end{center}
\end{table}

\subsection{Analysis of Model Component}

In this part, we demonstrate the effectiveness of each component proposed in our approach. As shown in Table. \ref{table:ab_module}, by leveraging the feature enhancement module (FEM) to boost low relevance caused by intra-class diversity and suppress noises derived from the local similarity of different categories of objects, we can improve the segmentation result from 61.3\% to 62.5\%. By employing the correlation reconstruction module (CRM) to supplement global context semantic information into correlation and keep support background information to correlations, we obtain extra 4.6\% performance gain. By employing the 2D residual decoder to refine the detail of objects, we can further improve the result to 67.4\%. 

Some qualitative segmentation examples on the PASCAL-$5^i$ dataset can be viewed in Fig \ref{fig:ab_module}. As can be seen, with the feature enhancement module (FEM), the network can remove false-positive segmentation and discover more object parts. For example, compared with base prediction, the segmentation of a person of the non-target class is removed in the third row, and the bird's unseen body is partially segmented in the first row. It shows that FEM can suppress noises and boost the low relevance of the region of the object. In addition, with the correlation reconstruction module (CRM), the model can capture a large object area and distinguish the background region in the query image. For example, compared with base+FEM, the non-target tower is removed in the second row, and most of the unseen body of the bird is distinguished in the first row. It indicates that the correlation reconstruction module can boost the encoder to learn a reliable and complete matching pattern. Finally, with the 2D residual encoder, our model can successfully refine the detail of object segmentation. For example, the marginal part of a bird's wing is segmented, and the outline of the plane is segmented more finely.

\begin{table}[tb]
	\renewcommand\arraystretch{1.15}
	\setlength{\tabcolsep}{1.5mm}
	\caption{Detailed ablation studies on FEM.}
	\label{table:ab_module_fem}
	\begin{center}
		\begin{tabular}{cccccc}
			\hline
			\textbf{Base} & \textbf{Cross-image} & \textbf{Intra-image} & \textbf{PASCAL-$\bm{5^i}$} &  {\textbf{COCO-$\bm{20^i}$}}\\
			\hline
			$\surd$  &    &         & 61.3   &  {34.2 }   \\
			$\surd$  &$\surd$         &    & 62.2   &  {36.6 }    \\
			$\surd$&      & $\surd$   & 61.8 &  {35.8}\\
			$ \surd$ & $\surd$     &  $\surd$        &  62.5  &   {38.0}\\

			\hline
		\end{tabular}
	\end{center}
\end{table}

\begin{figure*}
	\centering
	\includegraphics[height=10.5cm]{ 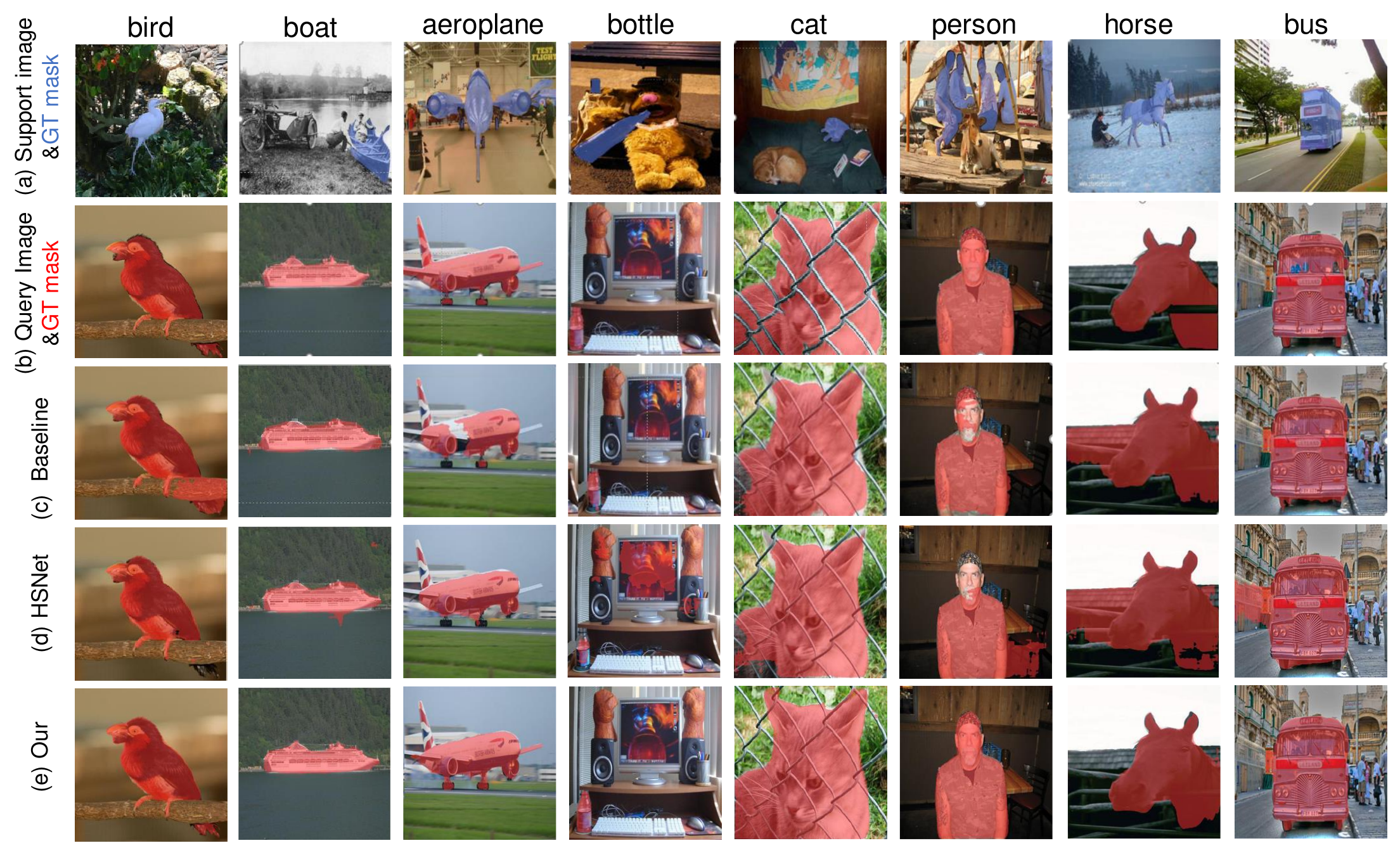}
	\caption{Example result on PASCAL-$5^i$ dataset for comparison with HSNet. From top to bottom, we show (a) support image with ground truth mask region in blue, (b) query image with ground truth mask region in red, (c) baseline prediction, (d) HSNet prediction, (e) our prediction.}
	\label{fig:ab_model}
\end{figure*}

 {More detailed ablation studies about our FEM are demonstrated in Table \ref{table:ab_module_fem}. As can be seen, each component of FEM contributes a lot on the challenging COCO-$20^i$. While the PASCAL-$5^i$ dataset is relatively simple and the noise can be alleviated to some extent by the encoder composed of 4D convolutions (convolution kernels are filters with certain abilities to adjust the correlation value adaptively), the baseline achieves a decent result of 61.3\%. Therefore, the improvements on PASCAL-$5^i$ are relatively minor. For the more complex COCO-$20^i$ datasets, the correlations derived from features contain too many noises that are beyond the adaptive adjustment ability of the encoder, leading to poor segmentation results. With the capability of suppressing the noises from complex images, our proposed FEM can effectively adjust the correlation and facilitate the encoder to learn better relationship matching. Specifically, with cross-image information communication at the spatial level, we can alleviate the noise caused by inter-class similarity and improve the result from 34.2\% to 36.6\% on COCO-$20^i$. With an intra-image channel attention module to enhance the support and query features representation, we can obtain another 1.4\% performance gain and reach 38.8\%. 
	
Table \ref{table:ab_module_crm} presents detailed ablation studies about our proposed CRM module. We can see that significant improvements are obtained on both PASCAL-$5^i$ and COCO-$20^i$. On PASCAL-$5^i$, with dense correlation to aggregate diversity semantic features for rich correlation information, we can improve the result to 63.1\%. Our background-keeping strategy brings 0.4\% performance gain compared to the background filtering operation adopted in  HSNet \cite{min2021hypercorrelation}. After leveraging global correlation to boost the encoder to learn more global visual pattern matching, we further improve the performance to 65.5\%. As can be seen, multi-scale guidance helps capture more complex self-similarity features and richer contextual semantic information, contributing to 0.5\% performance gain. Similar and more apparent improvements can be observed on COCO-$20^i$, which demonstrates the effectiveness of our proposed CRM module.
}

\begin{table}[htb]
	\renewcommand\arraystretch{1.2}
	\caption{Detailed ablation studies on CRM. DC: dense correlation, GC: global correlation.}
	\label{table:ab_module_crm}
	\setlength{\tabcolsep}{2.8mm}
	\begin{center}
		\begin{tabular}{lccc}
			\hline
			\textbf{Method} & \textbf{PASCAL-$\bm{5^i}$} &  {\textbf{COCO-$\bm{20^i}$}}\\
			\hline
			base   & 61.3  &  {34.2} \\
			base + DC (w/o background)   & 63.1  &  {36.1} \\
			base + DC   & 63.5  &  {36.8} \\
			base + DC  + GC (single-level)  & 65.0  &  {38.7} \\
			base + DC  + GC (multi-level)  & 65.5  &  {39.8} \\
			\hline
		\end{tabular}
	\end{center}
\end{table}

% \begin{figure}
% \centering
% \includegraphics[height=4cm]{ process.pdf}
% \caption{The training process comparison between HSNet and our model. Results are reported on fold 0 of PASCAL-$5^i$ using the ResNet50 backbone.}
% \label{fig:ab_process}
% \end{figure}

\subsection{Effectiveness in Constructing the Correlation} 
\label{subsec_ab}
In this part, we show experimental results and analysis of different correlation construction strategies to manifest the rationales behind our designs of the correlation reconstruction module. There are three alternatives for the correlation reconstruction: base correlation (BC), hyper-correlation construction (HC) used in HSNet and our proposed correlation reconstruction (CRM). Our experimental results in Table \ref{table:ab_CorConstruction} show that HC and CRM achieve significant improvement compared with BC, which proves that collecting abundant features to correlation is beneficial to improving the model's performance. Compared with HC, we notice that CRM improves performance from 64.0\% to 66.4\%, verifying that the global context semantic feature boosts the encoder to learn fine-grained correspondence relations between the query and the support images. The hyper-correlation used in HSNet is insufficient for the encoder to capture more complex and abundant visual patterns. Therefore, we design CRM to combine diverse semantic features and global context semantic information to construct superior correlation. Moreover, features endowed with global context are more robust to intra-class variations in CNN-based descriptors, which is beneficial for the encoder to segment objects accurately. \par 

\begin{table}[t]
\setlength{\tabcolsep}{8mm}
\renewcommand\arraystretch{1.4}
\caption{Ablation studies on the effect of correlation construction. The setting of experimental parameters is consistent with HSNet.}
\label{table:ab_CorConstruction}
\setlength{\tabcolsep}{4.8mm}
\begin{center}
\begin{tabular}{ccc}
\hline
\textbf{Method} & \textbf{Mean} \\
\hline
BC   & 61.3    \\
HC   & 64.0    \\
CRM   & 66.4    \\
\hline
\end{tabular}
\end{center}
\end{table}

\begin{figure*}[ht] %通栏
	\begin{minipage}[t]{0.32\linewidth} %调节两个子图左右间距
		\centering
		\includegraphics[width=4.5in]{ 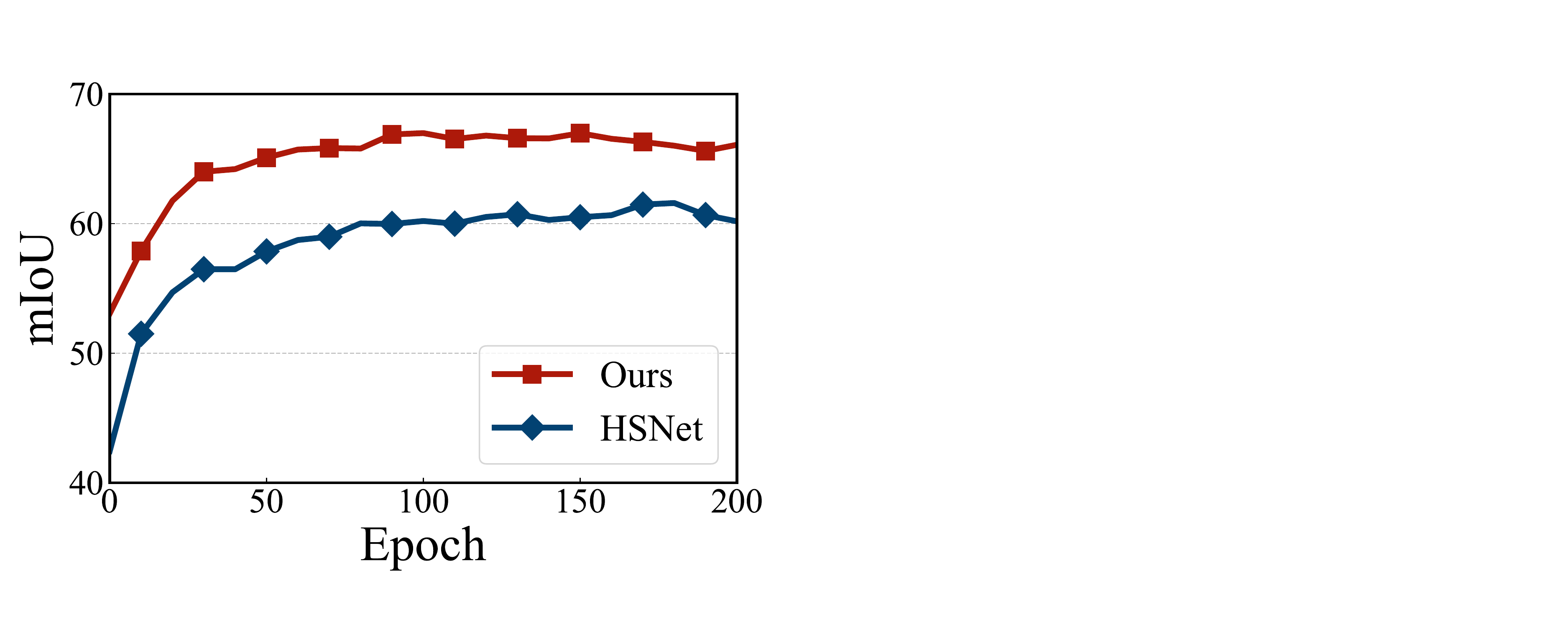} %调节单个子图大小
		\caption{The training process comparison between HSNet and our model. Results are reported on fold 0 of PASCAL-$5^i$.}
		% using the ResNet50 backbone
		\label{fig:ab_process}
	\end{minipage}%
	\hspace{1mm}
	\begin{minipage}[t]{0.32\linewidth}
		\centering
		\includegraphics[width=4.5in]{ 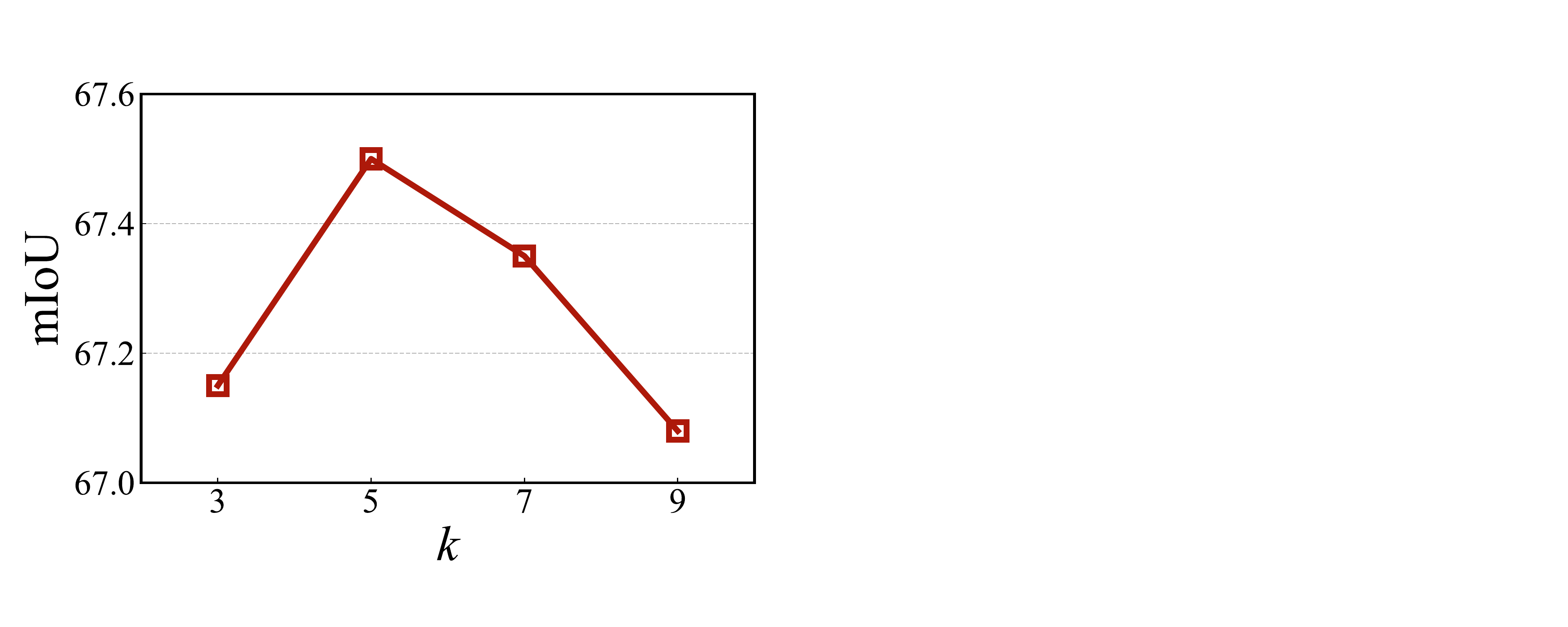}
		\caption{The parameter sensitivity of the neighbor region size $k$ in the local self-similarity generation of CRM. }
		\label{fig:ab_parameter_k}
	\end{minipage}
	\hspace{1mm}
	\begin{minipage}[t]{0.32\linewidth}
		\centering
		\includegraphics[width=4.5in]{ 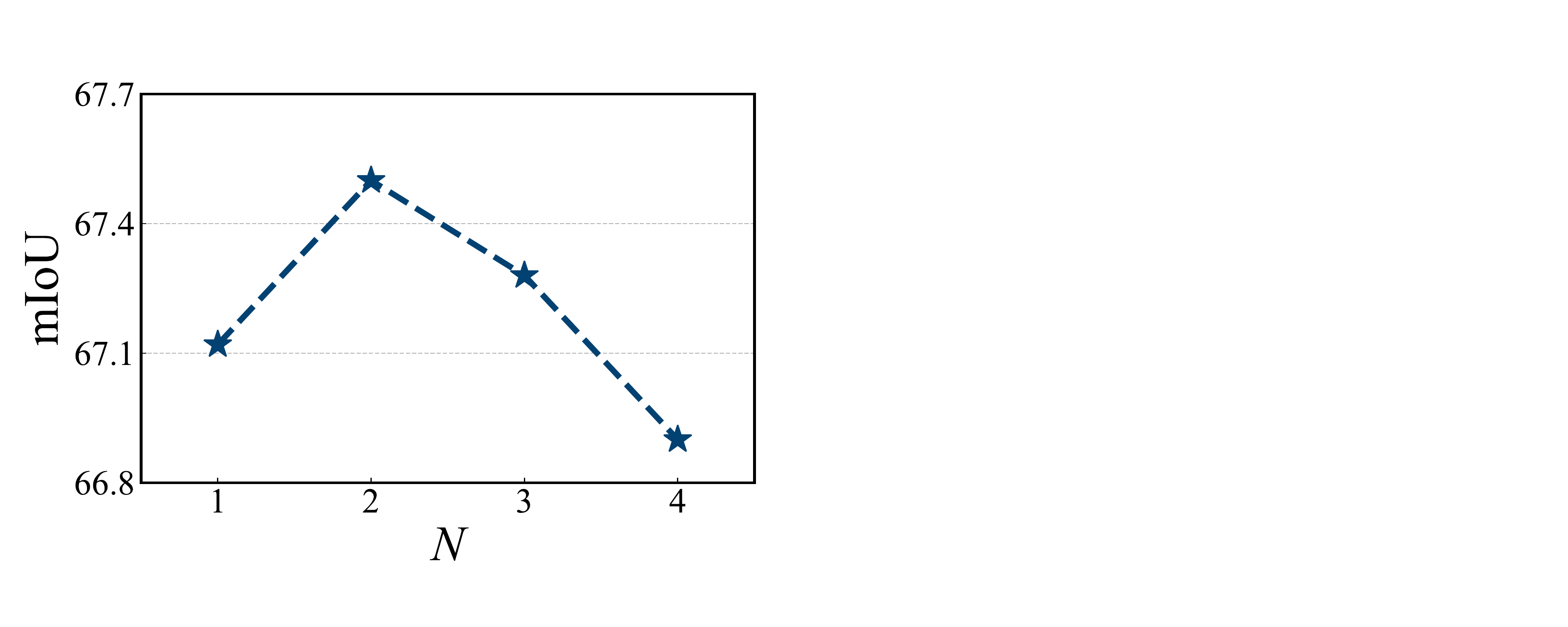}
		\caption{The parameter sensitivity of the convolution layer depth $N$ in the multi-scale similarity of CRM.}
		\label{fig:ab_parameter_N}
	\end{minipage}
\end{figure*}

\begin{table}[t]
	\renewcommand\arraystretch{1.2}
	\caption{Efficiency comparison with ResNet50 on PASCAL-$5^i$\cite{shaban2017one} in 1-shot setting.}
	\label{table:efficiency}
	
	\begin{center}
		\begin{tabular}{lcccc}
			\hline
			\textbf{Method} & \textbf{mIoU} & \textbf{Training} & \textbf{Inference} & \textbf{  Learnable params}\\
			\hline
			 {PFENet \cite{tian2020prior} } &  { 60.8 } &  { 24h    } &  { 52ms     } &  { 10.8M}       \\     
			 {RePRI \cite{boudiaf2021few}  } &  { 59.8 } &  { 17h    } &  { 72ms      } &  { 46.7M}           \\   
			 {CWT \cite{lu2021simpler}  } &  { 56.3 } &  { 10h    } &  { 232ms      } &  { 2.1M }    \\     
			 {MMNet \cite{wu2021learning}  } &  { 61.8 } &  { 64h    } &  { 128ms      } &  { 10.4M }   \\ 
			HSNet\cite{min2021hypercorrelation}  & 64.0& 54h &101ms &2.6M  \\
			Ours   &67.4& 33h &122ms &3.5M  \\
			\hline
			
		\end{tabular}
	\end{center}
\end{table}

We also display the comparison of the training process between HSNet and our model in Fig. \ref{fig:ab_process}. As we can see, our model converges faster and better than HSNet. One of the reasons for the poor effect of HSNet is that the noises in the correlation mislead the encoder to learn inappropriate relation matching. In addition, the lack of global context features makes the CNN-based encoder not robust. In contrast, our proposed FEM enhances low relevance caused by intra-class diversity and suppresses noises derived from the local similarity of different categories of objects. Moreover, our proposed CRM module removes unnecessary features in dense correlation and generates global context semantic features into correlations. \par

Some qualitative segmentation examples on the PASCAL-$5^i$ dataset can be viewed in Fig. \ref{fig:ab_model}. As can be seen, baseline prediction shows that part of the target object is not completely segmented (e.g., the tail of a bird in the first column, the body of an aeroplane in the third column and the head of a person in the sixth column) due to the correlation lack of global context semantic feature. HSNet employs extensive intermediate features to generate dense correlations, which contain information on diverse semantic features. However, we notice that prediction of HSNet exists plenty of false-positive segmentation (e.g., the bottle in the fourth column, the horse in the seventh column and the bus in the eighth column). As we mentioned above, hyper-correlation used in HSNet contains many noises and ignores the effect of background information in support images, misleading the encoder to learn inappropriate relation matching and treat background instances as objects of the target. Compared with these two methods, our model can capture more complete target objects. 

\subsection{Efficiency Comparison} 
\label{efficiency}

 {Table \ref{table:efficiency} demonstrates the efficiency comparison with previous state-of-art methods. As can be seen, compared to prototype-based methods \cite{tian2020prior,boudiaf2021few,lu2021simpler}, though the correspondence-based method of HSNet \cite{min2021hypercorrelation} significantly improves the performance, it takes much longer training time. In contrast, while our model further enhances the performance to 67.4\%, it converges faster than HSNet (previously shown in Fig. \ref{fig:ab_process}) and can reduce the training time from 54 to 33 hours. Since only 0.9M learnable parameters (ResNet50 contains about 25M) are introduced, our method slightly increases the inference time from 101 to 122 ms compared to HSNet.
}

\subsection{Parameter Analysis} 
\label{subsec_pa}
For self-similarity generation in the correlation reconstruction module, we conduct experiments to study the effect of the neighbor region size $k$. As shown in Fig. \ref{fig:ab_parameter_k}, we vary the neighbor region size $k$ over the range $\left \{3,5,7,9\right \}$. We notice that too large or small neighbor region size may not improve the results very much. We conjecture that a too small kernel size is insufficient to capture a wide range of global context feature, which is beneficial for the encoder to learn fine-grained correspondence relations. Meanwhile, a too large kernel size keeps too much background in query feature, hindering the model from segmenting the object accurately. In our experiments, we empirically set $k$ = 5. \par

% \begin{figure}[htbp]
%     \centering
%     \begin{subfigure}{0.48\textwidth}
%         \centering
%         \includegraphics[width=3]{ parameter_k.pdf}
%         \caption{The parameter sensitivity of the neighbor region size k in the local self-similarity generation of CRM. Results are reported on mean over all fold of PASCAL-$5^i$ using the ResNet50 backbone.}
%     \end{subfigure}
%     \qquad
%     \begin{subfigure}{0.48\textwidth}
%         \centering
%         \includegraphics[width=3]{ parameter_N.pdf}
%         \caption{The parameter sensitivity of the convolution layer depth N in the multi-scale similarity of CRM. Results are reported on mean over all fold of PASCAL-$5^i$ using the ResNet50 backbone.}
%     \end{subfigure}
% \end{figure}

% Results are reported on mean over all fold of PASCAL-$5^i$ using the ResNet50 backbone.
% \begin{figure}
% \centering
% \includegraphics[height=2.2cm]{ parameter_k.pdf}
% \caption{The parameter sensitivity of the neighbor region size k in the local self-similarity generation of CRM. Results are reported on mean over all fold of PASCAL-$5^i$ using the ResNet50 backbone.}
% \label{fig:ab_parameter_k}
% \end{figure}

% \begin{figure}
% \centering
% \includegraphics[height=2.2cm]{ parameter_N.pdf}
% \caption{The parameter sensitivity of the convolution layer depth N in the multi-scale similarity of CRM. Results are reported on mean over all fold of PASCAL-$5^i$ using the ResNet50 backbone.}
% \label{fig:ab_parameter_N}
% \end{figure}

For multi-scale similarity generation in the correlation reconstruction module, we conduct experiments to study the effect of the convolution layer depth $N$. As shown in Fig. \ref{fig:ab_parameter_N}, we vary the convolution layer depth $N$ over the range $\left \{1,2,3,4\right \}$. As we can see, we get the best performance when the convolution layer depth is 2. We guess that a too small number of the depth $N$ cannot extract complex context features. Meanwhile, a too large number of the depth $N$ contains many noises, which reduces the model's generalization ability. In our experiments, we empirically set $N$ = 2.

\section{Conclusion}
\label{conclusion}

In this work, we present a novel FECANet network for few-shot semantic segmentation task. Specifically, we proposed the feature enhancement module to filter noises in the correlation affected by local similarity and intra-class diversity by exchanging information between support and query features. The filtered correlation provides good guidance for the encoder to learn appropriate relation matching. To improve robustness to intra-class variations in a CNN-based encoder, we introduced a self-similarity method to integrate global context information into correlation. In addition, extensive experiments on PASCAL-$5^i$ and COCO-$20^i$ datasets demonstrated the superiority of our proposed model.

\end{document}